\documentclass[runningheads]{llncs}
\usepackage[T1]{fontenc}
\usepackage[misc]{ifsym}
\usepackage[table]{xcolor}
\usepackage{subfigure}
\usepackage{graphicx}
\usepackage{paralist}
\usepackage{color}
\usepackage{mfirstuc}
\usepackage{setspace}
\usepackage{graphicx}
\usepackage{etoolbox}
\usepackage[export]{adjustbox}
\usepackage{array}
\usepackage{float}
\usepackage{amsmath,amssymb,amsfonts}
\usepackage{cite}
\usepackage{color}
\usepackage{booktabs}
\usepackage{multicol,multirow}
\usepackage{caption}
\begin{document}
\title{Dynamic Multi-level Weighted Alignment Network for Zero-shot Sketch-based Image Retrieval}
\titlerunning{Dynamic Multi-level Weighted Alignment Network}
%

\author{Hanwen Su\orcidID{0009-0003-4335-5290} \and Ge Song \thanks{Corresponding author}\orcidID{0000-0002-2159-8203} \and Jiyan Wang\orcidID{0009-0000-4962-4496} \and Yuanbo Zhu\orcidID{0009-0004-1560-5966} }

\authorrunning{Su. et al.}

\institute{School of Computer and Electronic Information, Nanjing Normal University, Nanjing, 210023, CHN}

\maketitle

%
\begin{abstract}
The problem of \textit{zero-shot sketch-based image retrieval (ZS-SBIR)} has achieved increasing attention due to its wide applications, e.g. e-commerce. Despite progress made in this field, previous works suffer from using imbalanced samples of modalities and inconsistent low-quality information during training, resulting in sub-optimal performance. Therefore, in this paper, we introduce an approach called \textbf{Dynamic Multi-level Weighted Alignment Network for ZS-SBIR.} It consists of three components: (i) a Uni-modal Feature Extraction Module that includes a CLIP text encoder and a ViT for extracting textual and visual tokens, (ii) a Cross-modal Multi-level Weighting Module that produces an alignment weight list by the local and global aggregation blocks to measure the aligning quality of sketch and image samples, (iii) a Weighted Quadruplet Loss Module aiming to improve the balance of domains in the triplet loss. Experiments on three benchmark datasets, i.e., Sketchy, TU-Berlin, and QuickDraw, show our method delivers superior performances over the state-of-the-art ZS-SBIR methods.

\end{abstract}
\section{Introduction}
Sketch-based Image Retrieval (SBIR) aims to use a query hand-drawn sketch to retrieve images of the same category. With the explosive growth of content on the Internet, the SBIR has drawn increasing attention due to its capability of bringing convenience to potential users. However, in realistic scenarios, as databases expand to include new image categories, this leads to the study of extending the SBIR to the few-shot or even zero-shot settings (ZS-SBIR) \cite{ref_progress,ref_16,ref_domain,ref_eff,ref_sk3t,ref_tvt,ref_zse,shen2018zero,pang2019generalising,dey2019doodle,dutta2019semantically,liu2019semantic,gupta2022zero,dutta2020styleguide}. 
The key behind the ZS-SBIR problem is handling (i) the modality gap between sketch and image modalities, and (ii) the domain gap between seen and unseen categories. 

Previous works for ZS-SBIR such as \cite{ref_zse}, \cite{ref_tvt}, and \cite{liu2024zero} have made great progress. \cite{ref_zse} addressed the importance and the generalization capability of aligning the corresponding local patches for ZS-SBIR. \cite{ref_tvt} introduced a shared fusion ViT that can offer an extra fusion token for interaction with other tokens through the self-attention mechanism, resulting in a domain gap that is minimized more. Moreover, \cite{ref_zse} uses the triplet loss, which uses the imbalance samples of the modalities during training. \cite{liu2024zero} adjusts the margin parameter in the triplet loss adaptively, achieving better performance compared to methods using the original multi-modal triplet loss. However, we reveal that the alignment quality of the cross-modality data significantly impacts the zero-shot generalization of learned representation. 

Here, we introduce an approach that dynamically weights different sample pairs in a batch and creates cross-modal quadruplets during training. First, in a training batch, we weigh different pairs by calculating the alignment scores locally and globally, which represent the alignment quality for each pair. We leverage the tokens after cross-attention and calculate KL divergence to measure the alignment quality of class tokens and other tokens. The weights are used in the quadruplet we create during training.

 Secondly, we propose a weighted quadruplet loss to balance the number of samples from different modalities, inspired by \cite{ref_eff}. Unlike the original quadruplet loss that includes: a sketch anchor, positive image, negative image, and another negative image. We used the balanced numbers of sketch and image features by utilizing a negative sketch on top of the triplet loss. Besides, we dynamically weight the samples in the loss function to reduce the impact of low-quality alignment pairs.

\begin{figure*}[t]
    \centering
    \includegraphics[width=\linewidth]{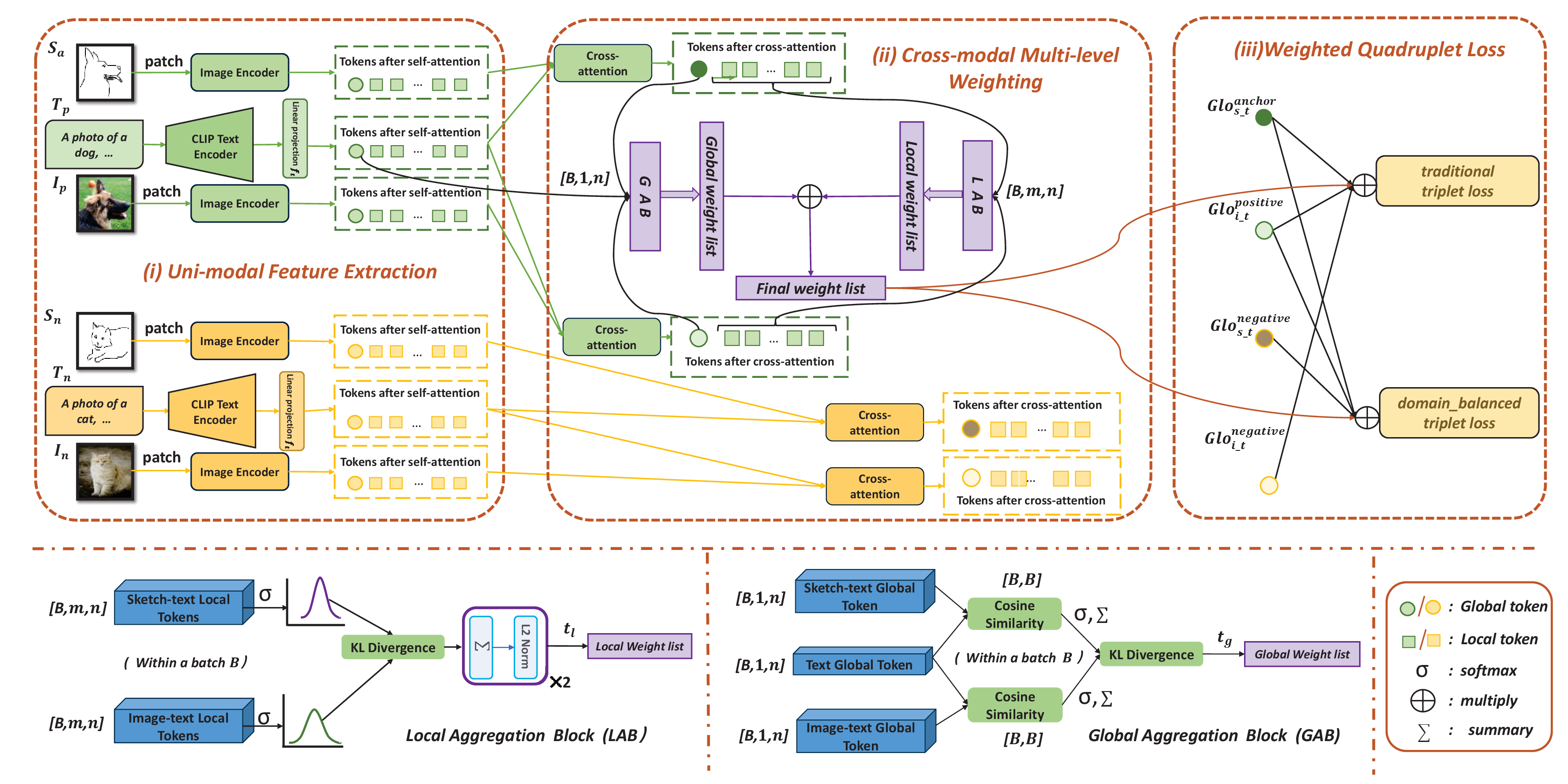}
    \caption{The overview of our proposed method. (i)Uni-modal Feature Extraction collects the text sketch and image into their corresponding backbone, a CLIP-16's text encoder, and a ViT for extracting sketch and image features. (ii) Cross-modal Multi-level Weighting takes tokens after cross-attention as its input and calculates the weights of each training sample in a batch. (iii) Finally weighted quadruplet loss is utilized for metric learning, which includes an anchor sketch, a negative sketch, a positive image, and a negative image.}
    \label{fig:overview}
\end{figure*}

Our main contributions can be summarized as:
\begin{compactitem}
\item We propose a dynamic weighting strategy that weights different pairs with their alignment quality, because higher quality improves the model's cross-modal zero-shot generalization capability.
\item We propose a weighted quadruplet loss that utilizes equal numbers of modals in the quadruplet loss and weights each sample with its alignment score.
\item Experiments conducted on 3 benchmark datasets of ZS-SBIR prove the effectiveness of our proposed approach.
\end{compactitem}

\section{Related Work}
\subsection{Zero-shot Sketch-Based Image Retrieval}
ZS-SBIR uses a query sketch to retrieve the images of the same category in the gallery. It aims to generate the knowledge learned from the seen training class into unseen categories. Introduced firstly by \cite{yelamarthi2018zero}, the ZS-SBIR aims to reduce the domain gap between sketches and photos by an image-to-image translation approximation. Subsequent work such as \cite{ref_progress,ref_eff,dutta2020styleguide,dutta2019semantically}
used CNNs as a backbones, then utilized projections of extracted features, and used the joint space. Besides, works such as \cite{ref_tvt,ref_zse} used ViT\cite{ref_16} as the backbone, achieving better results compared to CNN-based methods. Moreover, the teacher-student paradigm was employed by\cite{ref_domain,ref_eff} to distill the knowledge. In addition, a test-time training paradigm\cite{ref_sk3t} on the sketches in the test set was introduced to adapt to the test set distribution. \cite{ref_zse} addressed the importance and the generalization capability of aligning the corresponding local patches for ZS-SBIR. Recent method \cite{liu2024zero} proposed an adaptive relation-aware metric that improves the performance of triplet loss by adaptive adjusting the margin parameter during training. \cite{SainBCKXS23} leverage powerful generalisation ability of CLIP for ZS-SBIR with prompt learning. Similarly, \cite{ZhouL024} enhances the model by learning dataset-agnostic 'population distribution' of relevant modalities from CLIP. However, the above-mentioned methods got sub-optimal results, because they didn't consider the quality of pairing sketch-image during training.

\subsection{Language-assisted Vision Models}
Textual information contains much information and can be used to generalize to unseen domains. To utilize the zero-shot generalization ability, recent papers \cite{mao2023doubly,zhang2023prompt,naeem2023i2mvformer} have shown that using the textual information generated by LLMs with appropriate prompts can help the model learn complementary knowledge for vision tasks\cite{mao2023doubly,zhang2023prompt}, even for more fine-grained correspondence\cite{naeem2023i2mvformer}. Recently, \cite{dunlap2023using} has trained the retrieval model with text information, with the aim of extending to unseen domains and achieving a good retrieval result. In this paper, our base and full models leverage textual information to align the local patches of sketch and image.

\subsection{Retrieval Systems with Cross-modal Similarity}
Cross-modal similarity at the global and local levels reflects the aligning quality of a training pair, and it's a vital factor in a retrieval system's performance. Recently, a cross-domain video-text retrieval system \cite{hao2024uncertainty} analyzes the multi-granularity relationships between each target video and text to ensure the discriminability of target features. It inspires us to consider the similarity of each pairing sample at the global and local level, and then guide our model to focus more on high-quality training pairs. We found that the alignment quality of paired training data greatly impacts the model's zero-shot generalization ability. However, the previous ZS-SBIR methods take little account of this issue.

\section{Proposed Method}
In this section, we present our proposed method for ZS-SBIR. An overall diagram of the proposed approach is shown in Fig. \ref{fig:overview}. Each key module is detailed in the following.

\subsection{Uni-modal Feature Extraction}
For a given query sketch $S\in\mathbb{R}^{\textit{h}\times\textit{w}\times\textit{c}}$ and a gallery image $I\in\mathbb{R}^{\textit{h}\times\textit{w}\times\textit{c}}$, we use ViT\cite{ref_16} to extract features. More specifically, the ViT partition the input image into non-overlapping patches, then use a projection head to map them into vanilla tokens $S_v\in\mathbb{R}^{\textit{h}\times\textit{w}\times\textit{c}}$ and image feature $I_v\in\mathbb{R}^{\textit{h}\times\textit{w}\times\textit{c}}$.
We use the feature before the final MLP, which is semantically richer. We call this feature a global token $\mathit{[X_{[Glo]}]}$. As a consequence, the additional learnable global token $ \mathit{[X_{[Glo]}]} \in \mathbb{R}^{d}$ is inserted into the multi-head self-attention and MLP blocks to learn the global representation. For both sketch and image, the global tokens after multi-head attention can be attained by:
\begin{equation}
     \mathit{z_{0}} = [\mathit{X_{[Glo]}}; \mathit{X^1};...;\mathit{X^n}], 
\end{equation} 
\begin{equation}
     \mathit{z_{\ell}} = MSA(LN(z_{\ell-1}))+\mathit{z_{\ell-1}}, \mathit{l}=1...L,
\end{equation} 
\begin{equation}
    \mathit{z_{\ell}} = MLP(LN(z_{\ell}))+\mathit{z_{\ell}}, \mathit{l}=1...L,
\end{equation}
where $ \mathit{X} \in \{\mathit{S},\mathit{I}\}$,
LN represents the layer norm. L is the number of layers. And the residual connection is implemented. MSA is the multi-head self-attention:
\begin{equation}
        \mathit{S-Atten}  (\mathit{Q_{x}, K_{x}, V_{x}} ) = softmax(\frac{Q_{x}K_{x}^T}{\sqrt{d} } )V_{x},
\end{equation}

where the $\mathit{Q},\mathit{K},\mathit{V}$ are query, key, and value obtained by mapping the same token with three different linear projection heads $[\mathit{{W_{q}}}, \mathit{{W_{k}}}, \mathit{{W_{v}}}]$.
We use the Transformer model to extract its token features for a given text description \textit{D}. Following the \cite{ref_su}, the text descriptions \textit{D} are obtained by prompting the GPT-3\cite{ref_lang} with different training categories' names. The textual feature $\mathit{T}$ can be attained by:
\begin{equation}
    \mathit{T} = \mathit{f_{t}}(\mathit{E_{t}}(D)),
\end{equation}
where the $\mathit{E_{t}}$ stands for the text encoder, which is the CLIP\cite{ref_learning} text encoder corresponding to the ViT-B/16. $\mathit{f_{t}}$ is the linear layer for projecting the textual tokens to that of the same dimension of image and sketch features.

\subsection{Cross-modal Multi-level Weighting}
 In this module, we aim to calculate the weight of each sample in a training batch.
 First, we need to do cross-attention to obtain global and local tokens by:
     \begin{equation}
    \mathit{C-Atten_{x,t}}  (\mathit{Q_{t}, K_{x}, V_{x}} ) = softmax(\frac{Q_{t}K_{x}^T}{\sqrt{d} } )V_{x},
\end{equation} 
 
\textbf{(i) Local weight list.} In a training batch $\mathit{B}$, the Local Aggregation Block (LAB) takes the local tokens with the dimension of $[\mathit{m,n}]$ of sketch and image modalities as input. Then, we use softmax $\mathit{\sigma}$ to create two distributions $\mathit{D_s}, \mathit{D_i}$. Next, KL Divergence is used to measure the distance of $\mathit{D_s}$ and $\mathit{D_i}$. Finally, we summarize the output of KL Divergence with two stacks of summary and L2 Normalization to get the local alignment score whose size is $\mathit{B}$. Moreover, we produce the local weight list with a threshold $\mathit{t_l}$. It is the average of each sample's local alignment score within a batch $\mathit{B}$. If a sample's score is smaller than the threshold $\mathit{t_l}$, its weight will also be smaller.
The process can be formulated as:
\begin{equation}
 \mathit{dist\_kl\_l} = \mathit{KL(X_{local}^{i}, X_{local}^{s})}
\end{equation}
\begin{equation}
\mathit{local\_list} = \mathit{S(dist\_kl\_l)} 
\end{equation}
\begin{small}
\begin{equation}
\mathit{local\_list(i)}\\=\left\{\begin{array}{cc}
1, & \text { if } \mathit{local\_list(i)} \leq \mathit{t_l} \\
e^{local\_list(i)}-1, & \text { if } \mathit{local\_list(i)} > \mathit{t_l}
\end{array}\right.
\end{equation}
\end{small}
where $\mathit{X_{local}}$ stands for the local tokens. KL stands for KL Divergence, where we set sketch modality as the target. S stands for the stack, which is the combination of summary and L-2 Normalization.

\textbf{(ii) Global weight list.} We also measure the alignment quality globally. We utilize the text feature to act as the bridge between the sketch and image modality.  We assume that the greater cosine similarity of each pair demonstrates better alignment quality. The Global Aggregation Block (GAB) takes the text-sketch and text-image global feature to generate two matrices with the size of $\mathit{[B, B]}$. Then, we use softmax and summary to reduce the dimension of the matrix to the size of $\mathit{B}$. Similar to generating the local weight list, we use KL Divergence to measure the similarity. Finally, we produce the global weight list by a threshold $\mathit{t_g}$. The process can be formulated as:
 \begin{equation}
   \mathit{cos\_sim_{s/i}}=cos(X_{global}^{t}, X_{global}^{s/i})
\end{equation}

\begin{equation}
    \mathit{dist\_kl\_g} = \mathit{KL}(\sum (\sigma (\mathit{cos\_sim_{i}})),\sum (\sigma (\mathit{cos\_sim_{s}})) )
\end{equation}

\begin{small}
 \begin{equation}
\mathit{global\_list(i)}\\=\left\{\begin{array}{cc}
1, & \text { if } \mathit{global\_list(i)} \leq \mathit{t_g} \\
e^{global\_list(i)}-1, & \text { if } \mathit{global\_list(i)} > \mathit{t_g}
\end{array}\right.
\end{equation}     
\end{small}

where $\mathit{cos\_sim}_{s/i}$ stands for the cosine similarity between the textual tokens and text-sketch / text-image tokens.

Therefore, we can get the final alignment weight list by:
\begin{equation}
\mathit{fin\_list}=\mathit{local\_list}\times\mathit{globla\_list}
\end{equation} 

\subsection{Weighted Quadruplet Loss}
The original triplet loss for SBIR uses the triplet of \textless
$\mathit{{X_{global}^{s-anc}}}$, $\mathit{{X_{global}^{i-pos}}}$, $\mathit{{X_{global}^{i-neg}}}$\textgreater
to pull the positive pair away from the negative one. Moreover, quadruplet loss adds another negative image during training. Unlike the quadruplet loss, inspired by \cite{ref_eff}, we use a domain-balanced quadruplet, which minimizes the distance between sketches and images from the same semantic category while maximizing the distance between sketches and images from different categories in the target embedding space. It contains the quadruplet of \textless
$\mathit{{X_{global}^{s-anc}}}$, $\mathit{{X_{global}^{i-pos}}}$, $\mathit{{X_{global}^{i-neg}}}$, $\mathit{{X_{global}^{s-neg}}}$\textgreater
. Moreover, we multiply the weight of each training sample, using the final list attained in section 2.2. It can be broken down into two triplet losses (the original triplet loss $\mathit{L_{o}}$ and the domain-balanced triplet loss $\mathit{L_{d}}$):

\begin{small}
  \begin{equation}
    \mathit{L_{o}} = [ \left \| X_{global}^{s-anc}-X_{global}^{i-pos} \right \|-\left \| X_{global}^{s-anc}-X_{global}^{i-neg} \right \|+\alpha]_{+}
\end{equation}  
\end{small}
\begin{small}
  \begin{equation}
    \mathit{L_{d}} = [ \left \| X_{global}^{s-anc}-X_{global}^{i-pos} \right \|-\left \| X_{global}^{s-anc}-X_{global}^{s-neg} \right \|+\beta]_{+}
\end{equation}  
\end{small}
\begin{equation}
    \mathit{L_{wei\_qua}} = \sum_{i}^{B} \mathit{fin\_list}(i)\times(\mathit{L_{o}}(i) + \mathit{L_{d}}(i))
\end{equation}
where $[x]_{+}$ denotes max\{x,0\}, and $\| \cdot \|$  is the L-2 distance, $\alpha$ and $\beta$ are the margin parameters.

\subsection{Inference}
Notably, textual information is not used during inference, meaning only sketch and image features are required. We use the self-attention to extract the uni-modal features of the sketch and image modalities. Then, cross-attention is implemented on the uni-modal features, whose global tokens will be used to measure the similarity.

\section{Experiments}
\begin{table*}[h]
    \renewcommand{\arraystretch}{1.2}
\begin{center}
     \caption{Comparisons. “-": not reported, “\textdagger": approximate result. The best and second-best scores are colored in \textcolor{red}{red} and \textcolor{blue}{blue}.} 
    \centering
    \resizebox{\textwidth}{!}{
    \begin{tabular}{l r r r r r r r r r}
    \toprule[1pt]
    \multirow{2}[4]{*}{\textbf{Methods}} & \multicolumn{1}{l}{\multirow{2}[4]{*}{$\mathit{R^D}$}} & \multicolumn{2}{c}{\textbf{TU-Berlin}} & \multicolumn{2}{c}{\textbf{Sketchy-25}} & \multicolumn{2}{c}{\textbf{Sketchy-21}} & \multicolumn{2}{c}{\textbf{QuickDraw}} \\
\cmidrule{3-10}          &       & \multicolumn{1}{c}{mAP@all} & \multicolumn{1}{l}{Prec@100} & \multicolumn{1}{c}{mAP@all} & \multicolumn{1}{l}{Prec@100} & \multicolumn{1}{c}{mAP@200} & \multicolumn{1}{l}{Prec@200} & \multicolumn{1}{c}{mAP@all} & \multicolumn{1}{l}{Prec@200} \\
    \midrule
     ZSIH\cite{shen2018zero}  & 64    & 0.220  & 0.291  & 0.254  & 0.340  & \--  & \-- & \-- & \-- \\
    CC-DC\cite{pang2019generalising} & 256   & 0.247  & 0.392  & 0.311  & 0.468  & \-- & \-- & \-- & \-- \\
    DOODLE\cite{dey2019doodle} & 256   & 0.109  & \-- & 0.369  & \-- & \-- & \-- & 0.075  & 0.068  \\
    SEM-PCYC\cite{dutta2019semantically} & 64    & 0.297  & 0.426  & 0.349  & 0.463  & \-- & \-- & \-- & \-- \\
    SAKE\cite{liu2019semantic}  & 512   & 0.475  & 0.599  & 0.547  & 0.692  & 0.497  & 0.598  & 0.130  & 0.179  \\
    SketchGCN\cite{gupta2022zero} & 300   & 0.323  & 0.505  & 0.382  & 0.538  & \-- & \-- & \-- & \-- \\
    StyleGuide\cite{dutta2020styleguide} & 200   & 0.254  & 0.355  & 0.376  & 0.484  & 0.358  & 0.400  & \-- & \-- \\
    PDFD\cite{ref_progress}  & 512   & 0.483  & 0.600  & 0.661  & 0.781  & \-- & \-- & \-- & \-- \\
    ViT-Rec\cite{ref_16} & 512   & 0.438  & 0.578  & 0.483  & 0.637  & 0.416  & 0.522  & 0.115  & 0.127  \\
    DSN\cite{ref_domain}   & 512   & 0.484  & 0.591  & 0.583  & 0.704  & \-- & \-- & \-- & \-- \\
    SBTKNet\cite{ref_eff} & 512   & 0.480  & 0.608  & 0.553  & 0.698  & 0.502  & 0.596  & \-- & \-- \\
    Sketch3T\cite{ref_sk3t} & 512   & 0.507  & \-- & 0.575  & \-- & \-- & \-- & \-- & \-- \\
    TVT\cite{ref_tvt}   & 384   & 0.484  & 0.662  & 0.648  & 0.796  & 0.531  & 0.618  & 0.149  & \textcolor{red}{0.293}  \\
    ZSE-SBIR-Ret\cite{ref_zse} & 512   &0.569 & 0.637  & 0.736  & 0.808  & 0.504  & 0.602 & 0.142  & 0.202 \\
    ZS-SBIR-Adaptive\cite{liu2024zero} & 512 & 0.518 & 0.550 & \-- & \-- & \textcolor{blue}{0.532} & \textcolor{blue}{0.637} & \-- & \-- \\
    CMAAN-ATD~\cite{ref_su}  & 512   & \textcolor{blue}{0.617}  & \textcolor{blue}{0.682}  & \textcolor{blue}{0.763}  & \textcolor{blue}{0.819}  & 0.515  & 0.602  & \textdagger\textcolor{blue}{0.161}   & \textdagger{0.231}  \\ 
    \midrule
    Ours  & 512   & \textcolor{red}{0.623}  & \textcolor{red}{0.696}  & \textcolor{red}{0.787}  & \textcolor{red}{0.852}  & \textcolor{red}{0.572}  & \textcolor{red}{0.659}  & \textdagger\textcolor{red}{0.170}   & \textdagger\textcolor{blue}{0.254}  \\  
    \bottomrule[1pt]
    \end{tabular}}
    \label{tab:result}
\end{center}
\end{table*}

\begin{table}[t]
\caption{Ablation results. The best and second-best scores are bolded and underlined.}
\begin{center}
    \renewcommand{\arraystretch}{1.3}
    \begin{tabular}{ccc}
    \toprule[1pt]
        \textbf{method} & \textbf{Sketchy-25} & \textbf{Sketchy-21} \\ 
        \bottomrule[0.6pt]
        \textbf{metrics}  & \textbf{map@all, prec@100} & \textbf{map@200, prec@200} \\ 
        \textbf{base}  & 0.763, 0.819 & 0.515, 0.606  \\ 
        \textbf{w/o g+q}  & 0.770, 0.828 & 0.538, 0.628 \\ 
        \textbf{w/o g}  & \underline{0.786, 0.846} & \underline{0.566, 0.654}  \\ 
        \bottomrule[0.6pt]
        \textbf{ours} & \textbf{0.787, 0.852} & \textbf{0.572, 0.659}\\
        \bottomrule[1pt]
    \end{tabular}
\end{center}
\end{table}

\subsection{Implementation}
\textbf{Datasets.} There are three datasets commonly used for ZS-SBIR. \textit{Sketchy}\cite{ref_sketchy} consists of 75,471 sketches and 12,500 natural images from 125 classes. Sketchy-Ext\cite{ref_deep} is an extended version of the original Sketchy dataset, containing 125 categories. Moreover, an additional 60,502 photos are included. Sketchy-25 refers to a partition of 100 training classes and 25 testing classes. Sketchy-21\cite{ref_frame} refers to the version of 104/21 train/test classes, which selects classes that do not overlap with ImageNet\cite{ref_imagenet} categories as unseen classes. \textit{TU-Berlin}\cite{ref_tuberlin} contains 250 categories, each with 80 sketches. The 204,489 images provided by \cite{ref_deep} extends it. For the extended TU-Berlin dataset, following the \cite{liu2019semantic} we use 220 classes for training and test the remaining 30 classes. \textit{QuickDraw}\cite{dey2019doodle} is the largest SBIR dataset with 110 categories. It contains 330,000 sketches drawn by amateurs and 204,000 photos. The split of 80 classes for training and 30 for testing.

\textbf{Evaluation protocol.} Following the standard evaluation protocol, we test our model with mean average precision (mAP) and precision on top 100/200 (Prec@100/200).

\textbf{Implementation details.} We implement our model with the PyTorch toolkit. A sketch or image is scaled to 224*224. For network architecture, the self-attention consists of 12 layers, and ViT-B/16 is pre-trained on ImageNet-1K. The cross-attention is designed as one layer with 12 heads. For the optimizer, we choose Adam with weight decay 1e-2 and learning rate 5e-6.

\subsection{Results}

\begin{figure*}[t]
    \centering
    \includegraphics[width=1\linewidth]{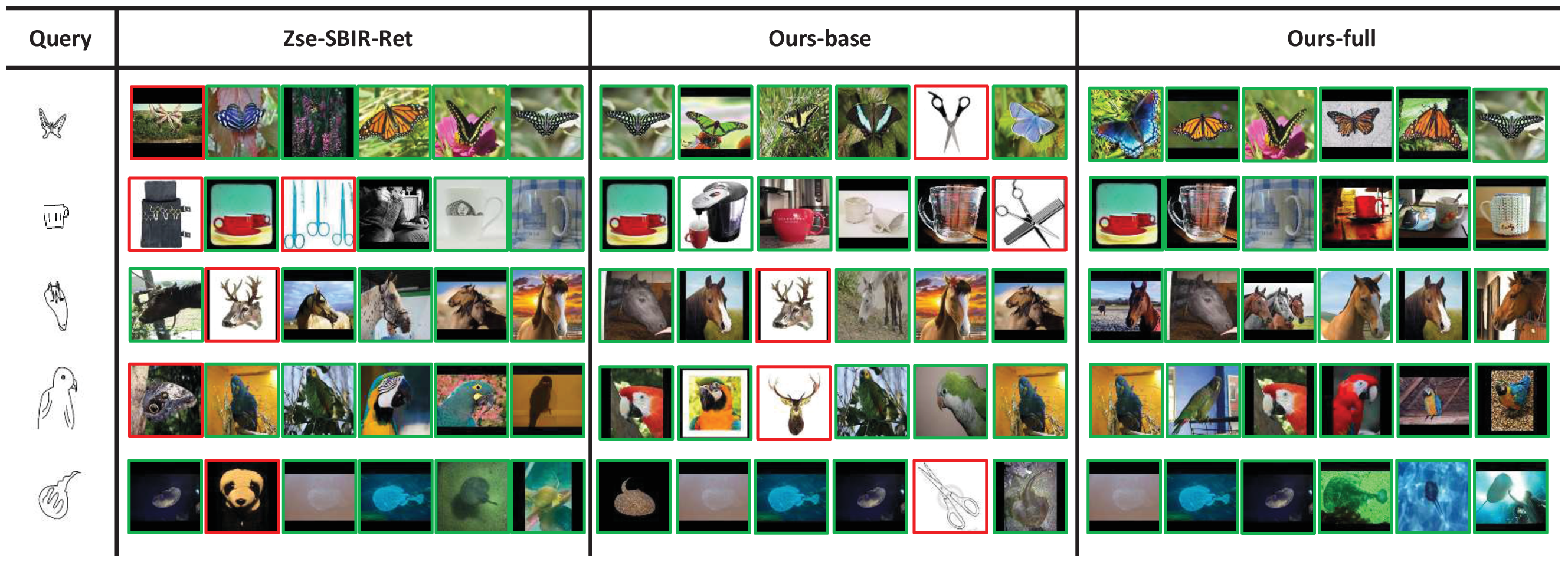}
    \caption{Exemplar comparison retrieval results for the given query sketches and the top 6 retrieved images. \textcolor{red}{Red} box denotes false positive, \textcolor{green}{Green} box denotes true positive.}
    \label{fig:qua}
\end{figure*}

\begin{figure*}[t]
    \centering
    \includegraphics[width=1\linewidth]{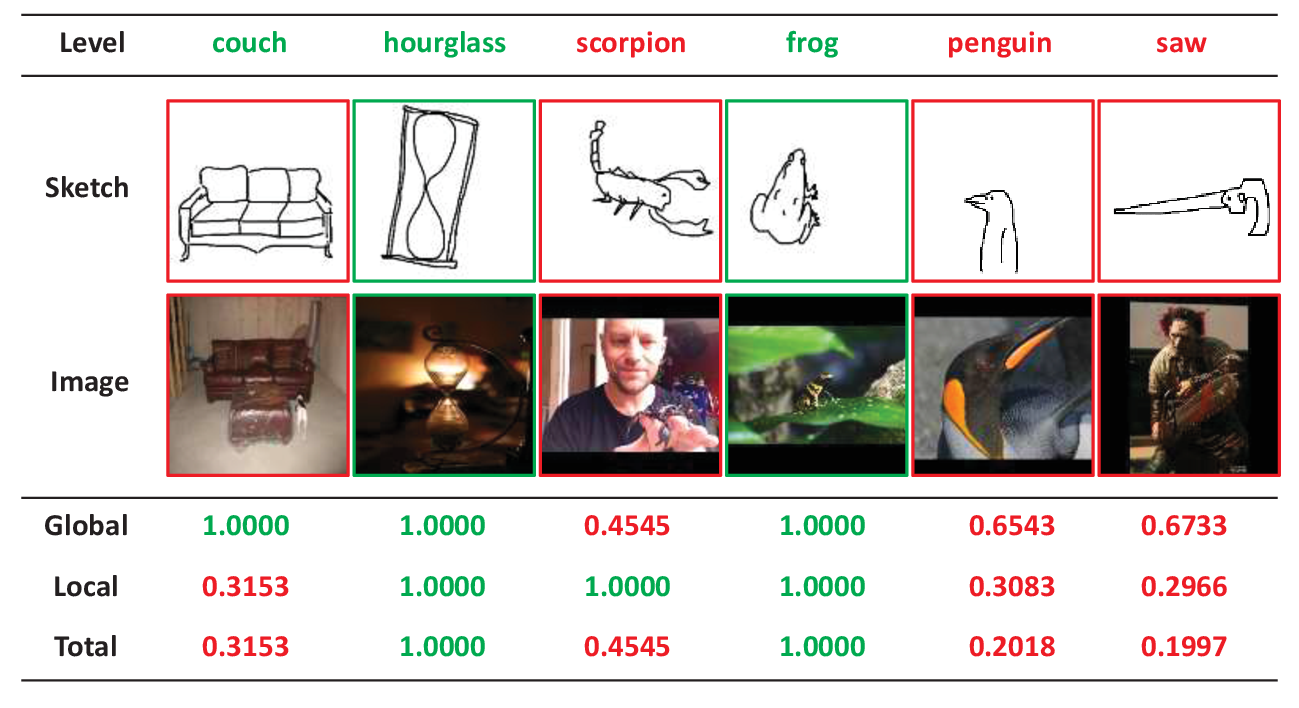}
    \caption{Results of multi-level weights reflect the similarities of sketch-image exemplar pairs during training. Closer to 1 means more similarity.}
    \label{fig:score}
\end{figure*}

From Table 1, we can observe that we achieve all top-3 results over all competitors. In addition, except for the QuickDraw dataset, our full model achieves all top-1. Due to the large data offered by QuickDraw, we test our method by randomly sampling 1/10 sketches to retrieve images and calculate the average by averaging ten results. 

\subsection{Further Study}

\textbf{Ablation Study.} We test the effectiveness of our method's modules on Sketchy-25 and Sketchy-21 datasets. (i) \textbf{base}: The method without cross-modal multi-level weighting uses triplet loss instead of quadruplet loss. (ii) \textbf{w/o g+q}: Adding only the local alignment score on the triplet loss on the base method. (iii) \textbf{w/o g}: Utilizing the weighted Quadruplet loss and multiplying it with the local alignment scores. (iv) \textbf{ours}: The full version of our method, which includes weighting the quadruplet loss with local and global alignment scores. We observe that without the weighted quadruplet loss, the model's performance has significantly deteriorated. Besides, removing the Cross-modal Multi-level Weighting module also leads to a decline in the model's performance.

\textbf{Qualitative analysis.} The top 6 retrieved candidates of sketch queries are shown in Fig.\ref{fig:qua}. We can observe that, compared with ZSE-SBIR-Ret\cite{ref_zse}, our models retrieve better. The CMAAN-ATD~\cite{ref_su} has the same structure as our model but ignores the quality of alignment,  resulting in the negative retrieval results on top. Moreover, the Ours model retrieves the most relevant candidates from the image gallery, proving the superiority of our proposed approach.

\textbf{Weight visualization.} During the first training epoch, we record multi-level similarity scores. The global and local scores are generated in the Global Aggregation Block(GAB) and Local Aggregation Block(LAB) in the Cross-modal Multi-level Weighting module. The total scores are obtained by multiplying global scores and the corresponding local scores. The results are shown in Fig.\ref{fig:score}.  We can observe that the sketch-image pairs with more complex backgrounds or with different angles and body gestures get lower points. While the samples with higher scores are thought to have higher aligning quality. It demonstrates the effectiveness of the proposed Cross-modal Multi-level Weighting module and the dynamic weighting strategy.

\section{Conclusion}
In this paper, we introduce a \textbf{Dynamic Multi-level Weighted Alignment Network for ZS-SBIR}. We dynamically calculate the similarity locally and globally to generate alignment scores for each training sample, which can measure the alignment quality of each training pair. Besides, we multiply the domain-balanced quadruplet loss by alignment scores, to reduce the impact of low-quality alignment paired samples. Experiments conducted on three benchmark datasets demonstrate the superiority of our approach. 
%
%
%

\bibliographystyle{splncs04}
\bibliography{reference}

\end{document}